\documentclass[11pt]{article}

\usepackage[final]{acl}
\usepackage{times}
\usepackage{latexsym}
\usepackage[T1]{fontenc}
\usepackage[utf8]{inputenc}
\usepackage{microtype}
\usepackage{url}
\usepackage{graphicx}
\usepackage{booktabs}
\usepackage{threeparttable}
\usepackage{amsmath}
\usepackage{amsthm}
\usepackage{amssymb}
\usepackage{amsfonts}
\usepackage{algorithm}
\usepackage{algorithmic}
\usepackage{multirow}
\usepackage{wrapfig}
\usepackage{pifont}
\usepackage{inconsolata}

\newtheorem{definition}{Definition}

\newcommand{\INPUT}{\textbf{Input:}}
\newcommand{\OUTPUT}{\textbf{Output:}}

\pdfmapfile{+cm.map}
\pdfmapfile{+symbols.map}
\pdfmapfile{+rsfs.map}

\title{Difficulty-Based Preference Data Selection by DPO Implicit Reward Gap}

\author{
\textbf{Xuan Qi}\textsuperscript{1*},
\textbf{Rongwu Xu}\textsuperscript{1*}, and 
\textbf{Zhijing Jin}\textsuperscript{2,3,4} \\
\textsuperscript{1}Paul G. Allen School of Computer Science \& Engineering, University of Washington\\
\textsuperscript{2}Max Planck Institute for Intelligent Systems, Tübingen, Germany \\
\textsuperscript{3}Jinesis Lab, University of Toronto \& Vector Institute \quad
\textsuperscript{4}EuroSafeAI \\
\texttt{\{xuanqi26,rongwuxu\}@cs.washington.edu},
\texttt{zjin@cs.toronto.edu}
}

\begin{document}

\maketitle

\begingroup
 \let\thefootnote\relax
\footnotetext{$^*$ co-first authors. The junior author is listed first.}
\endgroup

\begin{abstract}
Aligning large language models (LLMs) with human preferences is a critical challenge in AI research. While methods like Reinforcement Learning from Human Feedback (RLHF) and Direct Preference Optimization (DPO) are widely used, they often rely on large, costly preference datasets. The current work lacks methods for high-quality data selection specifically for preference data. In this work, we introduce a novel difficulty-based data selection strategy for preference datasets, grounded in the DPO implicit reward mechanism. By selecting preference data examples with smaller DPO implicit reward gaps, which are indicative of more challenging cases, we improve data efficiency and model alignment. Our approach consistently outperforms five strong baselines across multiple datasets and alignment tasks, achieving superior performance with only 10\% of the original data. This principled, efficient selection method offers a promising solution for scaling LLM alignment with limited resources. Code and data are available in \url{https://github.com/Difficulty-Based-Preference-Data-Select}.
\end{abstract}

\section{Introduction}
\label{sec:introduction}

Aligning large language models (LLMs) with human \emph{preferences} has emerged as one of the most critical challenges in recent AI research \cite{DBLP:conf/nips/ChristianoLBMLA17}. As LLMs demonstrate increasingly sophisticated capabilities across diverse domains~\cite{naveed2023comprehensive,thirunavukarasu2023large,qiu2025alita}, ensuring that their outputs align with human values and expectations becomes paramount for safe and beneficial deployment~\cite{yudkowsky2016ai,gabriel2020artificial,ouyang2022training}. Among the various alignment paradigms, Reinforcement Learning from Human Feedback (RLHF)~\cite{DBLP:conf/nips/ChristianoLBMLA17, ouyang2022training} has proven instrumental in fine-tuning state-of-the-art models. More recently, Direct Preference Optimization (DPO)~\cite{DBLP:conf/nips/RafailovSMMEF23} has gained significant traction as a computationally efficient alternative that bypasses explicit reward modeling while maintaining competitive performance. Central to the success of both algorithms is the quality of preference data that captures nuanced distinctions between desirable (e.g., helpful, honest) and undesirable (e.g., harmful, biased) model behaviors. However, as preference datasets scale to hundreds of thousands of examples (e.g., the widely used SHP dataset has 385K samples~\cite{pmlr-v162-ethayarajh22a}), the computational burden and potential inclusion of low-quality or redundant data points \cite{gao2024impact} necessitate data selection strategies. Effective curation of high-quality preference data not only \emph{reduces training costs} but also \emph{enhances model alignment} by focusing learning on the most informative preference signals.

Despite the critical importance of preference alignment, existing data selection methodologies for the LLM training pipeline \emph{predominantly} target instruction fine-tuning (IFT) datasets rather than preference datasets.\footnote{While IFT can be viewed as a form of \emph{behavior} alignment in the broad sense of AI alignment \cite{DBLP:conf/nips/ChristianoLBMLA17}, our focus here is on aligning models with explicit preferences.} Current approaches, including difficulty-based methods (filtering examples based on challenge level)~\cite{swayamdipta2020dataset, pleiss2020identifying, yuan2023self}, diversity-based techniques (selecting maximally heterogeneous subsets)~\cite{agarwal2021openbook, sorscher2022beyond, xie2023data}, and importance-based strategies (leveraging metrics to prioritize influential data points)~\cite{paul2021deep, mindermann2022prioritized, marion2023less}, are fundamentally designed for data in IFT. However, preference datasets have a fundamentally different structure: Each data point comprises an instruction paired with two responses, one chosen and one rejected, creating a comparative learning signal that requires specialized treatment. This structural distinction renders many IFT-oriented data selection algorithms inapplicable or suboptimal for working with preference datasets. Despite recent advances in preference alignment data filtering, such as SDPO~\cite{DBLP:journals/corr/abs-2502-09650}, these approaches have not adequately addressed the challenge of identifying high-quality preference data subsets, nor have they demonstrated sufficiently robust performance improvements. Consequently, the field currently lacks effective, theoretically grounded algorithms specifically designed for preference data selection, representing \emph{a significant gap in the LLM alignment toolkit}.

In response to this gap, we propose a novel \emph{difficulty-based} data selection method specifically designed for preference datasets. Our approach leverages the implicit reward mechanism inherent in the DPO algorithm~\cite{DBLP:conf/nips/RafailovSMMEF23} to quantify the difficulty of preference examples through the \textit{DPO implicit reward gap}, which is the difference between implicit rewards assigned to chosen and rejected responses. The core insight underlying our method is that preference examples with smaller reward gaps present greater learning challenges, as they represent boundary cases where the model exhibits uncertainty in distinguishing between preferred and rejected responses. This uncertainty manifests as higher gradient magnitudes during optimization, indicating greater learning potential due to amplified training signals at decision boundaries. Building on this theoretical foundation, we develop a systematic three-stage selection strategy: (1) computing DPO implicit reward gaps for all preference pairs using an aligned policy and its corresponding reference policy, (2) ranking examples by ascending reward gaps, and (3) selecting a subset where data points' reward gaps are under a certain threshold for downstream preference learning tasks. This principled approach ensures that selected examples provide maximum learning signal while maintaining computational efficiency.

To verify the effectiveness of our method, we carry out comprehensive empirical validations of our method across four preference datasets of diverse data types, including both human-annotated preferences ({SHP}~\cite{pmlr-v162-ethayarajh22a}) and synthetic datasets ({Skywork}~\cite{DBLP:journals/corr/abs-2410-18451}, {UltraFeedback}~\cite{DBLP:journals/corr/abs-2310-01377}, {RLHFlow}~\cite{RLHFlow2024}). Our evaluation covers two prevalent alignment tasks, reward model training and policy fine-tuning via DPO. Our approach is then benchmarked against five strong baselines. Results show that it consistently outperforms other data selection methods using the same amount of data. Furthermore, it even surpasses the models trained on the full dataset in over 67.5\% of cases,  achieving comparable or better performance while consuming only 10\% of the data. Additional analyses reveal: (1) Our method works robustly across different models for difficulty calculation; (2) The optimal data selection ratio falls between 10-15\%, and (3) Our approach remains effective even without length normalization. In total, these results establish our approach as both theoretically principled and practically effective for preference data selection of LLM alignment.

To summarize, our main contributions are as follows:
\begin{itemize}
    \item We propose a novel yet simple data selection method tailored for preference datasets, grounded in the theoretical framework of the DPO implicit reward mechanism to quantify sample difficulty.

    \item We provide a theoretical justification for our difficulty metric via gradient analysis, showing that smaller DPO implicit reward gaps correspond to larger gradient magnitudes, indicating higher learning potential.

    \item We perform extensive experiments on four diverse preference datasets and two alignment tasks, consistently achieving superior performance using only 10\% of the training data, outperforming five strong baselines and matching the performance of full-dataset training.

    \item We perform a comprehensive analysis of the method’s robustness under various difficulty computation models, data scaling regimes, and length normalization strategies, further identifying optimal selection ratios and demonstrating the method's robustness across different settings.
\end{itemize}

\section{Related Work}
\label{sec:related_work}

\subsection{Aligning LLM with Human Preferences}

Achieving alignment between LLMs and human preferences is a fundamental endeavor. A major advancement in this domain has been Reinforcement Learning from Human Feedback (RLHF)~\cite{DBLP:conf/nips/ChristianoLBMLA17, ouyang2022training, bai2022training}, which has played a pivotal role in the fine-tuning of leading LLMs such as GPT-4~\cite{openai2023gpt4}, Claude~\cite{anthropic2023claude}, and Gemini~\cite{team2023gemini} series models. The conventional RLHF approach involves training a reward model to evaluate the language model's outputs, followed by the application of reinforcement learning (RL) algorithms like Proximal Policy Optimization (PPO)~\cite{schulman2017proximal}, Trust Region Policy Optimization (TRPO)~\cite{schulman2015trust}, and others to fine-tune the model.

Despite its successes, PPO presents several challenges in alignment tasks, such as high complexity, instability, and inefficiency~\cite{casper2023open}. In response, studies have focused on improving the RLHF paradigm to achieve more robust alignment. Among these efforts, Direct Preference Optimization (DPO)~\cite{DBLP:conf/nips/RafailovSMMEF23} has emerged as a promising alternative, as it directly optimizes the model's policy based on human-annotated preference pairs, bypassing the need for a separate reward model. Other notable approaches include Identity Preference Optimization (IPO)~\cite{azar2024general}, Kahneman-Tversky Optimization (KTO)~\cite{ethayarajh2024kto}, and Simple Preference Optimization (SimPO)~\cite{meng2024simpo}. Our research builds upon the implicit reward mechanism in DPO, proposing an effective selection method for preference data that identifies high-quality preference pairs, ultimately enhancing model alignment.

\subsection{Data Selection for LLM Training}

Data selection plays a crucial role in the instruction fine-tuning (IFT) phase, as the quality and relevance of the IFT data significantly impact model performance~\cite{zhou2023lima, longpre2023flan}. Several strategies have been proposed to improve the efficiency and effectiveness of data selection, which can be coarsely categorized into three approaches: difficulty-based, diversity-based, and importance-based methods.

\paragraph{Difficulty-Based Methods} This line of methods focus on identifying and selecting data points that are challenging for the model to process or predict. For instance, \cite{swayamdipta2020dataset} use training dynamics to identify hard examples based on model confidence during training. \cite{pleiss2020identifying} leverage prediction uncertainty to select challenging examples that the model struggles with. More recently, \cite{yuan2023self} introduce a self-guided curriculum learning approach that progressively selects more difficult examples based on model performance. These methods typically leverage metrics such as perplexity or loss to quantify the difficulty of generating specific responses.  Our approach also belongs to this category. However, existing methods of this kind typically define the difficulty in the context of IFT data. In contrast, we propose the first difficulty-based data selection method specifically applied to preference datasets.

\paragraph{Diversity-based Methods} This line of methods prioritize selecting training data with a wide range of topics, styles, or contexts, thereby reducing redundancy and overlap between training examples. \cite{agarwal2021openbook} propose Core-Set selection methods that maximize coverage of the feature space. \cite{sorscher2022beyond} demonstrate that diversity-based selection can achieve comparable performance with significantly fewer training examples. \cite{xie2023data} introduce instruction diversity metrics specifically for IFT datasets. More recently, \cite{wu2023self} propose DiverseEvol, which uses a self-evolving mechanism to augment training datasets by selecting maximally dissimilar data points. The goal of these methods is to increase the diversity of the data, ensuring that the model learns from a broader spectrum of experiences.

\paragraph{Importance-based Methods} This line of methods assess the contribution of each data point to the overall training process, prioritizing those data points that have the greatest impact on model performance. \cite{paul2021deep} propose gradient-based importance sampling for neural network training. \cite{mindermann2022prioritized} introduce prioritized training on points that are likely to be forgotten, identifying influential examples through forgetting dynamics. \cite{marion2023less} develop LESS, which uses gradient-based influence estimation to select high-impact training examples. These methods often rely on metrics such as gradient magnitude, where data points that result in larger gradient updates are considered more important.

Relatively little attention has been paid to selecting preference data for LLM alignment. A notable exception is SDPO~\citep{DBLP:journals/corr/abs-2502-09650}, which prioritizes data using policy margin and reward-model uncertainty. In contrast, our method uses the DPO implicit reward gap to explicitly select small-gap boundary cases that provide stronger learning signals. To address this gap, we propose a simple yet effective algorithm that selects high-quality preference data based on its difficulty---quantified through the implicit reward gap in DPO. This provides a theoretically grounded approach to curating preference data for LLM alignment.

\section{Preliminary}
\label{sec:preliminary}

In this section, we provide necessary background on the Direct Preference Optimization (DPO) algorithm and the DPO implicit reward derived from it.

\subsection{Direct Preference Optimization}

Direct Preference Optimization (DPO)~\citep{DBLP:conf/nips/RafailovSMMEF23} provides an alternative to the traditional RLHF~\citep{DBLP:journals/corr/abs-2204-05862} paradigm by directly optimizing a policy using human-annotated preference pairs, eliminating the need for explicit reward model training. Given preference data $(x, y_w, y_l)$ where $x$ is the prompt, $y_w$ is the preferred (\underline{w}in) response, and $y_l$ is the rejected (\underline{l}ose) response, DPO loss tries to minimize:
\begin{equation}
\begin{aligned}
\ell_{\text{DPO}}(x, y_w, y_l; \theta)
&= - \log \sigma \Bigg( \beta \Bigg[
\log \frac{\pi_{\theta}(y_w | x)}{\pi_{\text{ref}}(y_w | x)} \\
&\quad - \log \frac{\pi_{\theta}(y_l | x)}{\pi_{\text{ref}}(y_l | x)}
\Bigg] \Bigg),
\end{aligned}
\end{equation}
where $\sigma$ denotes the logistic sigmoid function, $\beta$ is a hyperparameter that controls the strength of the preference signal, and $\pi_{\theta}$ represents the model's policy, parametrized by $\theta$. The function $\pi_{\text{ref}}$ refers to a reference model, which provides a baseline probability distribution over the responses. The term inside the logarithm represents the log-odds ratio between the chosen and rejected responses, weighted by $\beta$ to adjust the magnitude of the preference signal.

The key insight of DPO is that it implicitly defines a reward function without explicit reward modeling. The \textbf{DPO implicit reward} for any response $y$ given context $x$ is:
\begin{equation}
\label{eq:dpo-implicit}
\begin{aligned}
r_{\text{DPO}}(x, y)
&= \beta \log \frac{\pi_{\theta}(y | x)}{\pi_{\text{ref}}(y | x)} \\
&= \beta \sum_{t=1}^{|y|} \log
\frac{\pi_{\theta}(y_t | x, y_{<t})}
{\pi_{\text{ref}}(y_t | x, y_{<t})},
\end{aligned}
\end{equation}
where $|y|$ represents the length of response, and $y_{<t}$ represents the first t tokens of the response.

This implicit reward formulation exhibits several desirable properties that distinguish DPO from traditional RLHF approaches. The formulation naturally incorporates the reference model as a regularization term, preventing excessive deviation from the initial distribution while enabling token-wise decomposition for fine-grained optimization at each generation step~\citep{DBLP:conf/nips/RafailovSMMEF23,zeng2024token}. Unlike explicit reward models that suffer from distributional shift and require separate training phases, DPO's implicit reward remains inherently aligned with the policy throughout optimization, ensuring consistency and computational efficiency~\citep{DBLP:conf/nips/RafailovSMMEF23,park2024disentangling}. This direct encoding of human preferences into the optimization objective eliminates the need for reward model training while maintaining competitive performance with traditional RLHF methods~\citep{DBLP:conf/nips/RafailovSMMEF23}.

\section{Methodology}
\label{sec:method}

In this section, we introduce a novel method for selecting high-quality preference data based on their \textit{difficulty}, where difficulty is rigorously defined through the \textit{DPO implicit reward gap} (See Section \ref{sec:preliminary} for necessary background information on DPO). Specifically, we quantify the difficulty of a data point by measuring the gap between the DPO implicit rewards of the chosen and rejected responses. Our approach is grounded in the theoretical understanding that preference examples with smaller reward gaps present greater learning challenges and, consequently, offer higher potential for model improvement through optimization.

\subsection{Defining Difficulty of Preference Examples}

We define the difficulty of a training example as the gap between the DPO implicit rewards for the chosen and rejected responses. Let $x$ be the prompt, $y_{{w}}$ the chosen response, and $y_{{l}}$ the rejected response. The difficulty of a preference data example is quantified by the difference in the DPO implicit rewards between the chosen and rejected responses:
\begin{equation}
\Delta r_{\text{DPO}}(x,y_{{w}}, y_{{l}}) = r_{\text{DPO}}(x, y_{{w}}) - r_{\text{DPO}}(x, y_{{l}}),
\end{equation}
where $r_{\text{DPO}}(x, y)$ is the DPO implicit reward (see Equation \ref{eq:dpo-implicit}).
We hypothesize that preference examples with smaller reward gaps are more difficult for the model. A smaller gap implies greater uncertainty in distinguishing between the preferred and rejected responses, as the two are more similar in terms of the model's reward assignments.

\paragraph{Theoretical Justification for the Difficulty Metric} Our hypothesis that examples with smaller reward gaps present greater learning challenges can be justified through gradient analysis of the DPO optimization dynamics.

The DPO loss function for a single preference pair $(x, y_w, y_l)$ is given by:
\begin{equation}
\mathcal{L}_{\text{DPO}}(\theta) = -\log \sigma\left(\beta \Delta r_{\text{DPO}}\right),
\label{eq:dpo_loss}
\end{equation}
where $\sigma(\cdot)$ denotes the sigmoid function and $\beta > 0$ is the temperature parameter. For simplicity, in the following discussion we use $\Delta r_{\text{D}}$ to denote $\Delta r_{\text{DPO}}$.

Taking the gradient with respect to model parameters $\theta$, we obtain:
\begin{equation}
\frac{\partial \mathcal{L}_{\text{DPO}}}{\partial \theta} = -\beta \sigma(-\beta \Delta r_{\text{D}}) \frac{\partial \Delta r_{\text{D}}}{\partial \theta}.
\label{eq:gradient}
\end{equation}

The gradient magnitude is therefore:
\begin{equation}
\left\|\frac{\partial \mathcal{L}_{\text{DPO}}}{\partial \theta}\right\| = \beta \sigma(-\beta \Delta r_{\text{D}}) \left\|\frac{\partial \Delta r_{\text{D}}}{\partial \theta}\right\|.
\label{eq:gradient_magnitude}
\end{equation}

To analyze the relationship between reward gap and learning signal, we examine the sigmoid weighting factor $g(\Delta r_{\text{D}}) = \sigma(-\beta \Delta r_{\text{D}})$. This function achieves its maximum at:
\begin{equation}
\max_{\Delta r} g(\Delta r) = g(0) = \sigma(0) = \frac{1}{2},
\label{eq:max_sigmoid}
\end{equation}
which occurs precisely when $\Delta r_{\text{D}} = 0$.

For large positive reward gaps, we have:
\begin{equation}
\lim_{\Delta r_{\text{D}} \to +\infty} g(\Delta r_{\text{D}}) = \lim_{\Delta r_{\text{D}} \to +\infty} \sigma(-\beta \Delta r_{\text{D}}) = 0,
\label{eq:vanishing_positive}
\end{equation}
while for large negative gaps:
\begin{equation}
\lim_{\Delta r_{\text{D}} \to -\infty} g(\Delta r_{\text{D}}) = \lim_{\Delta r_{\text{D}} \to -\infty} \sigma(-\beta \Delta r_{\text{D}}) = 1.
\label{eq:vanishing_negative}
\end{equation}

However, in practice, negative reward gaps ($\Delta r_{\text{D}} < 0$) are undesirable as they indicate preference inversion. For well-aligned preference data where $\Delta r_{\text{D}} \geq 0$, the gradient magnitude in Equation~\eqref{eq:gradient_magnitude} is maximized when $\Delta r_{\text{D}}$ approaches zero, establishing that smaller reward gaps yield larger gradients and stronger learning signals.

Furthermore, the information-theoretic perspective supports this analysis. The uncertainty in preference distinction can be quantified by the entropy of the preference probability:
\begin{equation}
H(p) = -p \log p - (1-p) \log (1-p),
\label{eq:entropy}
\end{equation}
where $p = \sigma(\beta \Delta r_{\text{D}})$ represents the probability of preferring the chosen response. The entropy $H(p)$ is maximized when $p = 0.5$, corresponding to $\Delta r_{\text{D}} = 0$, indicating maximum uncertainty and thus maximum information content for learning.

This mathematical framework demonstrates that preference examples with smaller reward gaps $\Delta r_{\text{DPO}}$ provide both stronger optimization gradients and higher information content, thereby justifying their characterization as more difficult and valuable training examples.

\subsection{Data Selection Strategy}


Based on our theoretically grounded difficulty metric, the data selection strategy follows a systematic three-stage process: computing reward gaps (i.e., difficulty), ranking examples by difficulty, and selecting examples according to a predefined threshold.

\begin{itemize}
    \item \textbf{Stage 1: Difficulty Computation} For each preference data point $(x, y_{{w}}, y_{{l}}) \in D$  in the dataset, we compute the difficulty $\Delta r_{\text{DPO}}$ between the chosen and rejected responses using a DPO policy model $\pi_{\text{DPO}}$ and its reference policy model $\pi_{\text{ref}}$. It is crucial to note that the models used for difficulty calculation are typically \textit{different} from the target model to be trained. In a typical setup, $\pi_{\text{DPO}}$ is a pre-trained model that has already undergone preference alignment, while $\pi_{\text{ref}}$ is the corresponding model checkpoint before preference alignment, typically an instruction fine-tuned model. The selected data subset $D_{\text{select}}$ is then used to train a \emph{separate} target model, which may have a different architecture, scale, or initialization than the selection models. This \emph{decoupling} allows us to (1) leverage strong selector models to curate high-quality training data for potentially smaller or different target models, (2) repeatedly utilize the selected data subsets across various training paradigms, as the identification of high-quality preference data remains model-agnostic and independent of the downstream model being trained.
    \item \textbf{Stage 2: Difficulty Ranking} We rank all preference data points in ascending order according to their difficulty $\Delta r_{\text{DPO}}$. Examples with smaller gaps\footnote{We consider the numerical value of the gap rather than its absolute value.} are positioned higher in the ranking, as they present a greater learning potential.
    \item \textbf{Stage 3: Subset Selection} We select instances that either rank within the top $t$ percentile or exceed a predefined difficulty threshold $\tau$. Mathematically, the final selected dataset $D_{\text{select}}$ is defined as the subset of preference examples from $D$ for which the difficulty falls below a predefined threshold $\tau$:
\begin{equation}
\begin{aligned}
D_{\text{select}} = \{ &(x, y_{{w}}, y_{{l}}) \in D \mid \\
&\Delta r_{\text{DPO}}(x, y_{{w}}, y_{{l}}) \leq \tau \}.
\end{aligned}
\end{equation}

The threshold $\tau$ can be determined either as a fixed value selected through preliminary experiments (see Section \ref{analysis:scaling-effects} for exploration on the optimal ratio)  or dynamically based on a desired selection ratio $\rho \in (0, 1)$ by taking the $\rho$-quantile of the computed reward gaps:
\begin{equation}
\begin{aligned}
\tau = \text{quantile}\big(&\{\Delta r_{\text{DPO}}(x, y_{{w}}, y_{{l}}) : \\
&(x, y_{{w}}, y_{{l}}) \in D\}, \rho\big).
\end{aligned}
\end{equation}
\end{itemize}

\begin{figure}[ht]
    \centering
\includegraphics[width=\linewidth]{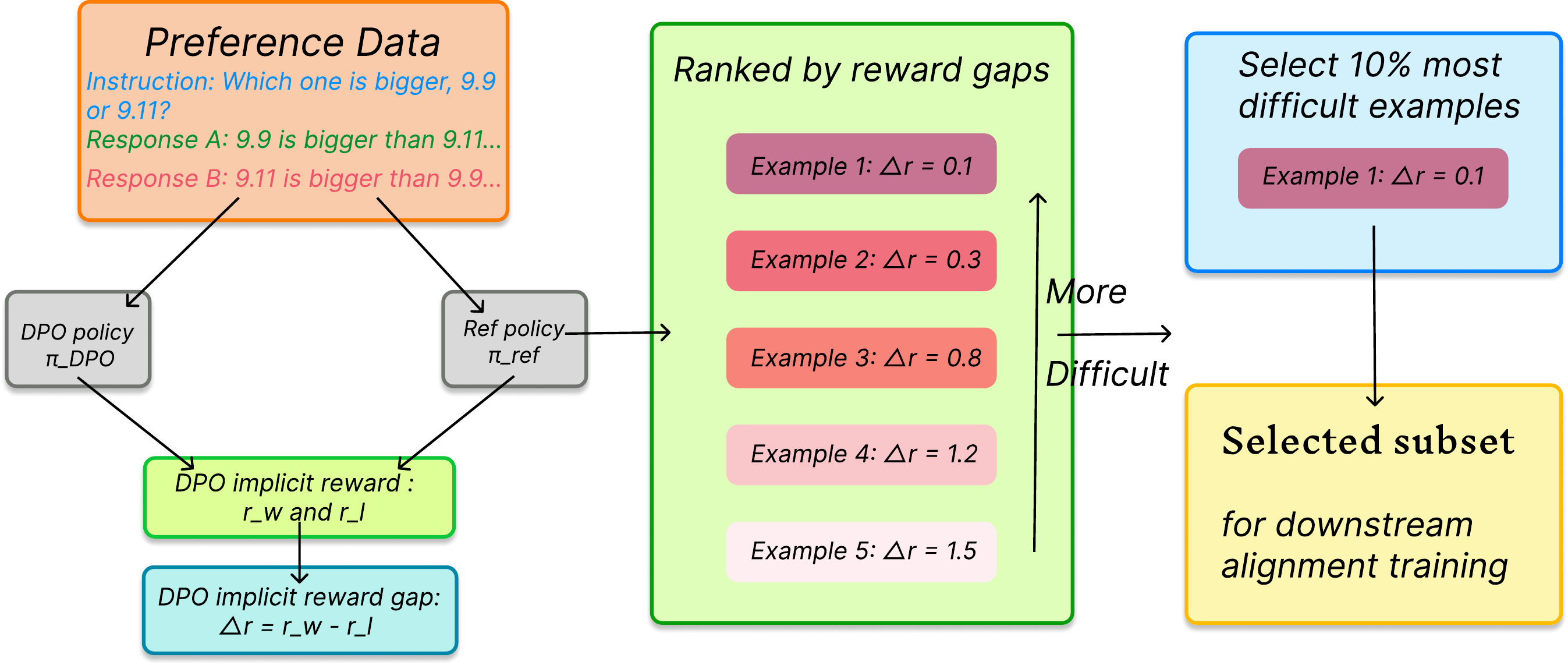}
    \caption{Illustration of our preference data selection pipeline.}
    \label{fig:data-pipeline}
\end{figure}

To provide a clearer understanding of our method, Algorithm~\ref{alg:difficulty_based_selection} presents the complete workflow of our difficulty-based preference data selection method. This methodology prioritizes the most difficult examples. The threshold $\tau$ can be adjusted based on empirical results to fine-tune the selection process, balancing between data quality and quantity according to model capacity and the specific alignment task requirements. Figure~\ref{fig:data-pipeline} illustrates our data selection pipeline.

\begin{algorithm}[t]
\caption{Difficulty-Based Preference Data Selection}
\label{alg:difficulty_based_selection}
\small
\begin{algorithmic}[1]
\STATE \INPUT\ Preference dataset $D = \{(x_i, y_{w,i}, y_{l,i})\}_{i=1}^N$
\STATE \INPUT\ DPO policy model $\pi_{\text{DPO}}$, reference model $\pi_{\text{ref}}$
\STATE \INPUT\ Selection ratio $\rho \in (0, 1)$ (or threshold $\tau$)
\STATE \OUTPUT\ Selected preference subset $D_{\text{select}}$

\STATE {\small \textcolor{gray}{\# Stage 1: Reward Gap Computation}}
\STATE Initialize reward gap list $\Delta R = []$
\FOR{each preference pair $(x_i, y_{w,i}, y_{l,i}) \in D$}
    \STATE Compute $r_{\text{DPO}}(x_i,y_{w,i}) = \beta \log \frac{\pi_{\text{DPO}}(y_{w,i} | x_i)}{\pi_{\text{ref}}(y_{w,i} | x_i)}$
    \STATE Compute $r_{\text{DPO}}(x_i,y_{l,i}) = \beta \log \frac{\pi_{\text{DPO}}(y_{l,i} | x_i)}{\pi_{\text{ref}}(y_{l,i} | x_i)}$
    \STATE Calculate reward gap $\Delta r_i = r_{\text{DPO}}(x_i, y_{w,i}) - r_{\text{DPO}}(x_i, y_{l,i})$
    \STATE Append $\Delta r_i$ to $\Delta R$
\ENDFOR

\STATE {\small \textcolor{gray}{\# Stage 2: Difficulty Ranking}}
\STATE Sort indices by ascending reward gaps: $\text{indices} = \text{argsort}(\Delta R)$

\STATE {\small \textcolor{gray}{\# Stage 3: Subset Selection}}
\IF{selection ratio $\rho$ is provided}
    \STATE $\tau = \text{quantile}(\Delta R, \rho)$
\ENDIF
\STATE $D_{\text{select}} = \{(x_i, y_{w,i}, y_{l,i}) \in D : \Delta r_i \leq \tau\}$

\RETURN $D_{\text{select}}$
\end{algorithmic}
\end{algorithm}

\paragraph{Computational Cost Analysis}
Given the practical implementation considerations for large-scale datasets, we hope to understand the efficiency trade-offs in our difficulty-based selection approach. The reward gap computation stage dominates the computational complexity of our data selection method. For a dataset $D$ with $|D| = N$ preference pairs, the computational cost can be analyzed as follows:

\begin{itemize}
    \item \textbf{Forward Pass Complexity:} Each reward gap computation requires forward passes through both the DPO policy model $\pi_{\text{DPO}}$ and reference model $\pi_{\text{ref}}$ for two responses per preference pair. This results in $4N$ forward passes with complexity $\mathcal{O}(NC_{\text{forward}})$, where $C_{\text{forward}}$ represents the cost of a single forward pass.
    
    \item \textbf{Ranking Complexity:} The ranking stage requires sorting $N$ reward gaps, contributing $\mathcal{O}(N \log N)$ comparison operations.
    
    \item \textbf{Selection Complexity:} The final selection stage operates in $\mathcal{O}(N)$ time for threshold-based selection or $\mathcal{O}(N)$ for quantile-based selection.
\end{itemize}

The overall computational complexity is $\mathcal{O}(N C_{\text{forward}} + N \log N)$, which is dominated by the forward pass computation. Importantly, this cost is incurred only once during the preprocessing stage and does not affect the training efficiency of the downstream alignment process. Moreover, the computational overhead is amortized across the entire training process, as the selected high-quality subset typically leads to faster convergence and better final performance. Appendix~\ref{appendix:computation-complexity} provides a comparison with the baseline selection methods used in Section~\ref{sec:experiments}.

For practical implementation, the method can be parallelized across GPUs, and the computed reward gaps can be cached for multiple experiments with different selection thresholds $\tau$, further improving computational efficiency.

\section{Experiments}
\label{sec:experiments}

In this section, we conduct comprehensive experiments to evaluate the effectiveness of our proposed difficulty-based data selection method across multiple preference datasets for aligning LLMs. Our experimental evaluation encompasses two critical tasks: (1) reward model training (\texttt{RM}) and (2) policy alignment using DPO (\texttt{DPO}). Through systematic comparison against several state-of-the-art data selection baselines\footnote{Due to the limited number of methods specifically designed for preference data selection, we adapt several approaches that originally target IFT data selection.}, we demonstrate that our method consistently achieves superior performance compared to other methods.

\subsection{Experimental Setup}

\paragraph{Datasets}

\begin{table}[h]
\centering
\begin{threeparttable}
\begin{tabular}{lcc}
\toprule
\textbf{Dataset} & \textbf{Size} & \textbf{Type} \\
\midrule
\texttt{SHP} & 385K & Human \\
\texttt{Skywork}& 77K & Synthetic \\
\texttt{UltraFeedback} & 61K & Synthetic \\
\texttt{RLHFlow} & 100K & Synthetic  \\
\bottomrule
\end{tabular}
\end{threeparttable}
\caption{Statistics of preference datasets used in our experiments. Size: number of preference pairs, Type: whether preferences are human-annotated or synthetically generated.}
\label{tab:dataset_statistics}
\end{table}

We evaluate our method on four representative preference datasets that span both human-annotated preferences and synthetic ones, including human-annotated preference dataset \texttt{SHP} \citep{pmlr-v162-ethayarajh22a} and synthetic preference datasets {Skywork-Reward-Preference-80K-v0.2} (\texttt{Skywork}) \citep{DBLP:journals/corr/abs-2410-18451}, {ultrafeedback-binarized} (\texttt{UltraFeedback}) \citep{DBLP:journals/corr/abs-2310-01377}, {RLHFlow-pair-data-v2-80K-wsafety} (\texttt{RLHFlow})~\cite{RLHFlow2024}.  These datasets vary in scale and annotation quality, providing a comprehensive testbed for our approach. Table~\ref{tab:dataset_statistics} presents detailed statistics for each dataset.
Synthetic preferences are typically derived through automated proxy evaluation systems, such as rule-based scoring. For instance, in the \texttt{UltraFeedback} dataset, multiple model responses to a given instruction are automatically scored across dimensions, with the highest and lowest scoring responses forming the preferred and rejected examples, respectively.

\paragraph{Models}

For difficulty calculation in our experiments, we use the \texttt{LLaMA3-iterative-DPO-final} model~\cite{xiong2024iterative, dong2024rlhf} as the DPO policy model and its supervised fine-tuning (SFT) checkpoint, \texttt{LLaMA3-SFT}, trained from \texttt{Llama-3-8B}~\citep{dubey2024llama}, as the reference model.

For the \texttt{RM} task, we pick \texttt{gemma-2-2b-it}~\citep{gemma_2024} as the base model and follow the implementation outlined in RLHFlow~\citep{dong2024rlhf} to train a standard Bradley-Terry reward model~\citep{Bradley1952RankAO}. For the \texttt{DPO} task, we use {Tulu3-Llama3.1-8B-SFT} (\texttt{Tulu3-SFT}) \citep{lambert2024tulu} as the base model for DPO and follow the implementation outlined in OpenRLHF~\citep{hu2024openrlhf} to fine-tune the model.

All experiments are performed using NVIDIA 80GB A100 or H100 GPUs.

\begin{table*}[h]
\centering
\fontsize{9}{9}\selectfont
\begin{threeparttable}
\resizebox{\textwidth}{!}{%
\begin{tabular}{clcccccccc}
\toprule
\textbf{Dataset} & \textbf{Dimension} & \textbf{Ours} & \textbf{Full Set} & \textbf{Random} & \textbf{ZIP$^\dag$} & \textbf{DiverseEvol$^\dag$} & \textbf{SDPO} \\
\cmidrule{1-8}
\multirow{5}{*}{\texttt{SHP}} & Chat & \underline{0.8073} & \textbf{0.8198} & 0.7874 & 0.7933 & 0.7791 & 0.7860\\
 & Chat-Hard & \textbf{0.6342} & \underline{0.6039} & 0.5155 & 0.5734 & 0.5364 &  0.5593\\
 & Safety & \textbf{0.8059} & 0.7906 & 0.7698 & \underline{0.7926} & 0.7864 & 0.7802\\
 & Reasoning & 0.5531 & 0.5624 & 0.5592 & \textbf{0.5764} & \underline{0.5631} & 0.5508\\
 & Total & \textbf{0.7056} & 0.7008 & 0.6882 & \underline{0.7012} & 0.6954 & 0.6923\\
\midrule
\multirow{5}{*}{\texttt{Skywork}} & Chat & \textbf{0.8798} & 0.8603 & 0.8659 & \underline{0.8705} & 0.8611 & 0.8654 \\
 & Chat-Hard & \textbf{0.7785} & 0.6885 & 0.6425 & 0.6845 & \underline{0.7054} & 0.6930 \\
 & Safety & \textbf{0.8446} & 0.8014 & 0.7783 & 0.7926 & \underline{0.8029} & 0.7993 \\
 & Reasoning & 0.6138 & \textbf{0.8350} & 0.6339 & 0.6283 & \underline{0.6419} & 0.6328 \\
 & Total & \underline{0.7588} & \textbf{0.7812} & 0.7189 & 0.7283 & 0.7359 & 0.7306 \\
\midrule
\multirow{5}{*}{\texttt{UltraFeedback}} & Chat & \textbf{0.8098} & 0.7946 & 0.7844 & \underline{0.7961} & 0.7958 & 0.7954 \\
 & Chat-Hard & \textbf{0.6425} & 0.6029 & 0.5983 & \underline{0.6327} & 0.6041 & 0.6217 \\
 & Safety & \textbf{0.7632} & 0.7416 & 0.7384 & 0.7493 & 0.7299 & \underline{0.7544} \\
 & Reasoning & 0.6904 & \textbf{0.7056} & 0.6886 & \underline{0.6971} & 0.6781 & 0.6701 \\
 & Total & \underline{0.7327} & \textbf{0.7391} & 0.7018 & 0.7288 & 0.7063 & 0.7193 \\
\midrule
\multirow{5}{*}{\texttt{RLHFlow}} & Chat & \textbf{0.8062} & 0.7291 & 0.7152 & \underline{0.7983} & 0.7855 & 0.7961 \\
 & Chat-Hard & 0.7098 & \underline{0.7127} & 0.6938 & \textbf{0.7142} & 0.7024 & 0.7090 \\
 & Safety & \textbf{0.8219} & 0.8081 & 0.7914 & \underline{0.8093} & 0.7956 & 0.7942 \\
 & Reasoning & 0.6985 & \textbf{0.7723} & \underline{0.7558} & 0.7265 & 0.7038 & 0.6957 \\
 & Total & 0.7524 & \underline{0.7562} & 0.7392 & \textbf{0.7614} & 0.7493 & 0.7515 \\
\bottomrule
\end{tabular}}
\end{threeparttable}
\caption{Task \texttt{RM}: Performance of reward models trained across data selection methods, evaluated on RewardBench's different splits: \textbf{Chat}, \textbf{Chat-Hard}, \textbf{Safety}, and \textbf{Reasoning} with \textbf{Total} being the average score. \textbf{Bold} indicates the highest score in each row, and \underline{underlined} indicates the second-highest score. $^\dag$ denotes methods adapted from IFT-oriented data selection.}
\label{tab:rewardbench_results}
\end{table*}

\begin{table*}[h]
\centering
\fontsize{9}{9}\selectfont
\begin{threeparttable}
\resizebox{\textwidth}{!}{%
\begin{tabular}{clccccccccc}
\toprule
\textbf{Dataset} & \textbf{Dimension} & \textbf{Tulu3-SFT} & \textbf{Ours} & \textbf{Full Set} & \textbf{Random} & \textbf{ZIP$^\dag$} & \textbf{DiverseEvol$^\dag$} & \textbf{SDPO} \\
\midrule
\multirow{2}{*}{\texttt{SHP}} & LCWR  & 2.57 & \textbf{17.92} & \underline{17.84} & 16.58 & 17.22 & 16.98 & 16.58 \\
 & WR  & 2.16 & \textbf{16.74} & \underline{16.52} & 15.49 & 16.03 & 15.77 & 15.96 \\
\midrule
\multirow{2}{*}{\texttt{Skywork}} & LCWR & 2.57 & \textbf{20.56} & 18.13 & \underline{19.60} & 18.74 & 17.75 & 17.46 \\
 & WR & 2.16  & \textbf{19.38} & 17.54 & 18.57 & \underline{18.96} & 18.33 & 18.56 \\
\midrule
\multirow{2}{*}{\texttt{UltraFeedback}} & LCWR  & 2.57 & \underline{18.41} & \textbf{18.44} & 17.53 & 17.83 & 17.20 & 16.69 \\
 & WR  & 2.16 & \textbf{19.52} & 16.82 & \underline{17.49} & 16.77 & 16.59 & 15.74 \\
\midrule
\multirow{2}{*}{\texttt{RLHFlow}} & LCWR & 2.57  & \textbf{19.85} & \underline{18.74} & 18.57 & 18.34 & 17.52 & 18.09 \\
 & WR & 2.16  & \textbf{19.44} & 17.93 & \underline{18.13} & 18.06 & 16.73 & 17.83 \\
\bottomrule
\end{tabular}}
\end{threeparttable}
\caption{Task \texttt{DPO}: Performance of DPO fine-tuned models across data selection methods, evaluated on Alpaca 2.0 Eval's two metrics: \textbf{WR} (Win Rate, model wins vs. reference) and \textbf{LCWR} (Length-controlled WR, mitigating length bias). The remaining experimental settings are identical to the experiment on Task \texttt{RM}. \textbf{Bold} indicates the best performance, and \underline{underlined} indicates the second-best performance.}
\label{tab:dpo_results}
\end{table*}

\paragraph{Baselines}
To benchmark our method, We compare against the following strong baselines: Full Set,  Random, ZIP~\citep{yin2024entropy}$^\dag$\footnote{$^\dag$ denotes methods adapted from IFT-oriented data selection.}, DiverseEvol~\citep{wu2023self}$^\dag$ and SDPO~\citep{DBLP:journals/corr/abs-2502-09650}. And the specific details of the baseline methods can be found in Appendix~\ref{app:experiment}.

In the experiments, to ensure a fair comparison, we use the full original dataset as the ``baseline of all baselines'' (Full Set). For all data selection methods, only 10\% of the data is selected for training.

\paragraph{Evaluation Metrics}
We assess model performance for the two alignment tasks using two separate metrics:

\begin{itemize}
    \item \textbf{Accuracy on RewardBench (for \texttt{RM}):}
    For reward model evaluation, we report the accuracy on the RewardBench~\citep{DBLP:journals/corr/abs-2403-13787}. Accuracy is defined as the proportion of test instances where the reward model assigns a higher score to the chosen response.
    \item \textbf{GPT-4o Win Rate (for \texttt{DPO}):} For DPO-tuned models, we evaluate on the AlpacaEval 2.0 benchmark~\citep{alpaca_eval}. Each model generates responses to a standard set of instructions and is compared to a default baseline using GPT-4o~\citep{DBLP:journals/corr/abs-2410-21276} as the judge\footnote{We adopt the specific configurations and prompts from AlpacaEval 2.0 as detailed in \url{https://github.com/tatsu-lab/alpaca_eval?tab=readme-ov-file\#alpacaeval-20}.}. The win rate is computed as the percentage of test cases where the model's response is rated better than the Full Set baseline.
\end{itemize}

\subsection{Results}

\paragraph{\texttt{RM}: Reward Model Training}
We train reward models using datasets selected by different data selection strategies, along with the Full Set baseline, and evaluate them on RewardBench across four preference datasets. Table \ref{tab:rewardbench_results} summarizes the performance across four dimensions (Chat, Chat-Hard, Safety, and Reasoning) and an aggregated score (Total). Our method consistently outperforms baseline data selection approaches across multiple datasets, often achieving performance comparable to models trained on the full dataset despite using significantly fewer examples. When compared with other baselines excluding the Full Set, our method demonstrates superior performance on the complete RewardBench dataset, achieving optimal results in 75\% of the evaluation cases. Across the various dimensions of RewardBench assessment, our approach outperforms all baseline methods in 69\% of scenarios, significantly surpassing alternative methodologies. Notably, our approach demonstrates remarkable data efficiency, it even surpasses the models trained on the full dataset in over 67.5\% of cases,  achieving comparable or better performance while consuming only 10\% of the data.

The method exhibits robust performance across diverse data characteristics, from synthetic scenarios to human-annotated discussions, suggesting that our difficulty-based selection principle captures fundamental aspects of preference learning that generalize beyond specific data-generation procedures. Comparison with SDPO, which is the only method specifically designed for data selection in the preference alignment domain, reveals that our reward gap approach, which directly targets learning potential, provides superior outcomes compared to margin-based selection strategies, supporting our theoretical analysis.

\paragraph{\texttt{DPO}: Policy Alignment Using DPO}
We fine-tune models using DPO with different strategies across various datasets and evaluate performance using GPT-4o as a judge on the AlpacaEval 2.0 benchmark. Table \ref{tab:dpo_results} presents the results, which further validate that our data selection strategy yields more informative and high-quality preference subsets. Our proposed methodology consistently outperforms all other baseline approaches across various experimental settings. When compared against the Full Set baseline, our method demonstrates superior performance in 88\% of cases, exceeding the capabilities of models trained using DPO on the complete dataset.

The results demonstrate improved or comparable performance relative to models trained on full datasets while consistently outperforming other baselines with the same data budget. The DPO experiments corroborate the data efficiency advantages observed in reward model training, confirming that our difficulty-based selection approach effectively identifies the most valuable training examples for policy alignment across different optimization frameworks.

Compared to the \texttt{RM} task, our method demonstrates more pronounced advantages in the \texttt{DPO} task with the selected dataset. This can be attributed to our approach using the DPO implicit reward gap for data selection, which aligns the defined difficulty more consistently with the difficulty of each data point in DPO training, thereby achieving superior performance.

Overall, the dataset selected by our method maintains high performance levels across both tasks, outperforming other baselines, and in many settings, achieving comparable results to the Full Set. These findings validate the superiority of our data selection methodology.

\section{Additional Experiments and Analysis}
\label{sec:analysis}

In Appendix \ref{app:additional-human-data}, we report results for additional review-stage experiments further confirm wall-clock efficiency, rule out a length-only explanation, and show competitive results on Anthropic Helpful-Harmless data.

In Appendix \ref{app:analysis}, we show additional analysis on model choice, selection ratio, response length normalization, and the selected examples' statistical characteristics. These results show that our method is robust across different selector model pairs, works best around a 10-15\% selection ratio, and remains effective without length normalization.

\section{Conclusion}
\label{sec:conclusion}

In this work, we introduce a novel difficulty-based data selection method for preference datasets, grounded in the DPO implicit reward mechanism. By focusing on preference examples with smaller reward gaps, our method identifies the most challenging data points, which offer higher learning potential for model alignment. Through extensive experiments across multiple preference datasets, we demonstrated that our approach consistently outperforms existing data selection strategies, achieving superior performance while using only a fraction of the data. The method's robustness and efficiency across various datasets and alignment tasks underline its potential for enhancing the training of large language models. Future work may explore further refinements to the selection strategy, as well as its integration into other paradigms beyond DPO.

\section*{Limitations}

While our data selection method demonstrates strong performance across multiple datasets and alignment tasks, several limitations should be acknowledged. First, our approach requires access to a pre-trained DPO policy model and its corresponding reference model for difficulty computation, which may not always be available in resource-constrained settings. Second, our method's effectiveness depends on the quality of the selector models used for difficulty calculation; if these models are poorly aligned or biased, the selected data may inherit these limitations. Third, while we demonstrate robustness across different selector models, the optimal selection ratio (10-15\%) may vary depending on the specific dataset characteristics and downstream tasks. Finally, our current approach focuses on single-stage selection and does not explore adaptive or iterative selection strategies that might further improve data efficiency.

\section*{Ethics Statement}

This work addresses data selection for aligning LLMs with human preferences, which has important ethical implications. Our method aims to improve the efficiency of preference alignment by identifying high-quality training data, potentially reducing computational costs and environmental impact. However, we acknowledge several ethical considerations:

\paragraph{Data Quality and Bias} Our selection method relies on models trained on existing preference datasets, which may contain biases or reflect problematic human preferences. Careful curation and auditing of training data remain essential to ensure aligned models promote beneficial outcomes.

\paragraph{Transparency and Reproducibility} We provide detailed methodology and experimental results to facilitate reproducibility. The effectiveness of our method depends on access to aligned LLMs, which may not always be publicly available.

\paragraph{Deployment Considerations} Models trained using our selected data should be carefully evaluated for safety, fairness, and alignment with human values before deployment. Data selection is one component of responsible AI development, but comprehensive evaluation and monitoring remain critical.

We commit to following ethical guidelines in AI research and encourage the community to consider these factors when applying our method.

\section*{Usage of AI Assistants}

We only use AI assistants for language polishing. The authors take charge of all technical contents.

\section*{Acknowledgments}

We would like to extend our gratitude to our colleagues, the reviewers, area chairs, and other ACL program committee members for providing valuable feedback on our work.

This material is based in part upon work supported by the German Federal Ministry of Education and Research (BMBF): Tübingen AI Center, FKZ: 01IS18039B; by the Machine Learning Cluster of Excellence, EXC number 2064/1 – Project number 390727645; by Schmidt Sciences SAFE-AI Grant; by the Frontier Model Forum and AI Safety Fund; by Coefficient Giving;
by
the Canadian AI Safety Institute Research Program
at CIFAR;
by the Canadian AI Safety Institute Research Program
at CIFAR through a Catalyst Award;
by the Survival and Flourishing Fund;
and
by the Cooperative AI Foundation.
The usage of OpenAI credits is largely supported by the Tübingen AI Center and Schmidt Sciences.
Resources used in preparing this research project were provided, in part, by the Province of Ontario, the Government of Canada through CIFAR, and companies sponsoring the Vector Institute.

\bibliography{references}

\appendix

\section{Computational Complexity Comparison with Other Baseline Methods}
\label{appendix:computation-complexity}

We provide a comparison of the computational complexity between our proposed method and other baseline approaches. Table~\ref{tab:complexity-comparison} summarizes these comparisons, where $N$ represents the total number of data samples in the dataset, $C$ denotes the cost of a basic computational operation (such as a forward pass or compression calculation), and $T$ represents the number of iterations or training steps where applicable.

Our method achieves an overall complexity of $\mathcal{O}(NC + N \log N)$, primarily involving forward passes for reward gap computation and sorting operations. This computational cost is incurred only once during preprocessing and leads to faster convergence and better final performance in the downstream alignment process.

\begin{table}[h]
\centering
\begin{threeparttable}
\begin{tabular}{lc}
\toprule
Method & Time Complexity \\
\midrule
Ours & $\mathcal{O}(NC + N \log N)$ \\
ZIP$^\dag$ & $\mathcal{O}(N + TN \log N)$ \\
DiverseEvol$^\dag$ & $\mathcal{O}(TN^2)$ \\
SDPO & $\mathcal{O}(TC + NC)$ \\
\bottomrule
\end{tabular}
\end{threeparttable}
\caption{Computational complexity comparison with simplified notation.}
\label{tab:complexity-comparison}
\end{table}

Our method's efficiency stems from its streamlined approach that requires only a single preprocessing stage, without the need for multiple model training iterations (as in SDPO) or quadratic comparison operations between samples (as in DiverseEvol). This makes our approach particularly suitable for large-scale datasets where computational efficiency is paramount.

We further measure end-to-end wall-clock time on \texttt{Skywork}. Under the same downstream DPO recipe, our pipeline, including reward-gap computation, subset selection, and DPO training on the selected 10\% subset, finishes in 83 minutes, compared with 126 minutes for full-data DPO training. This corresponds to a 34.1\% wall-clock reduction.

\begin{table*}[h]
\centering
\begin{threeparttable}
\begin{tabular}{lccc}
\toprule
\textbf{Pipeline} & \textbf{DPO Data} & \textbf{Included Steps} & \textbf{Time} \\
\midrule
Ours & 10\% & Gap computation + selection + DPO & 83 min \\
Full Set & 100\% & Full-data DPO training & 126 min \\
\bottomrule
\end{tabular}
\end{threeparttable}
\caption{End-to-end wall-clock comparison on \texttt{Skywork}. Both runs use the same downstream DPO training recipe; lower is better.}
\label{tab:wallclock_efficiency}
\end{table*}

\section{Further Experimental Details on Baselines}
\label{app:experiment}

To benchmark our method, we compare it against the following strong baselines. Here are the detailed descriptions of those methods:

\begin{itemize}
     \item \textbf{Full Set:} The original dataset without any filtering or subsampling, representing an upper bound in terms of available data volume, and serves as a reference point to assess the performance of \emph{all} data selection methods.
    \item \textbf{Random:} Random means choosing a random subset of the dataset. This baseline controls for the effect of subset size and allows us to isolate the contribution of informed data selection strategies.
    \item \textbf{ZIP~\citep{yin2024entropy}$^\dag$\footnote{$^\dag$ denotes methods adapted from IFT-oriented data selection.}:} ZIP is a model-free data selection method grounded in the principle that data with lower compression ratios, e.g., text that is harder to compress, typically contains more unique patterns, diverse vocabulary, and complex structures that tend to contain more effective information. ZIP identifies a subset of training data by iteratively minimizing the overall compression ratio using a multi-stage greedy algorithm.
    \item \textbf{DiverseEvol~\citep{wu2023self}$^\dag$:} A diversity-driven data selection method that leverages a self-evolving mechanism to augment the training dataset iteratively. At each step, DiverseEvol selects data points that are maximally dissimilar from those already chosen, based on the model's current embedding space. This is implemented via a K-Center-based sampling strategy.
    \item \textbf{SDPO~\citep{DBLP:journals/corr/abs-2502-09650}:} SDPO uses a model-based data selection method that selects training samples by prioritizing those with large policy margin and low reward model uncertainty, aiming to mitigate gradient instability and ensure more consistent policy updates. \emph{Crucially, SDPO differs from our approach in its focus on policy margin and uncertainty rather than reward gap difficulty}. While SDPO aims to mitigate gradient instability through margin-based selection, our method specifically targets the most challenging examples that provide maximum learning potential through small reward gaps.
\end{itemize}

\section{Additional Review-Stage Experiments}
\label{app:additional-human-data}

\subsection{Efficiency via Wall-Clock Time}

To address the concern that our method’s practical efficiency gains may be offset for many practitioners, and that prior experiments did not quantify wall-clock savings relative to SDPO’s iterative training, we report a direct efficiency comparison against full-dataset training in Table \ref{tab:wallclock_efficiency}. The result indicate training on the full set is 51.8$\%$ slower compared to on the subset selected through our method.

\subsection{Additional Results on Human Preference Dataset}

To evaluate whether the proposed selection strategy transfers beyond the primary benchmarks, we conduct an additional experiment on the Anthropic Helpful-Harmless (HHH) human preference dataset~\citep{bai2022training}. We use the same selection pipeline and downstream reward model training protocol as in the main \texttt{RM} experiments. Results are reported in Table \ref{tab:hhh_rewardbench}.
These results indicate that our method remains highly competitive on HHH, suggesting that the proposed selection strategy also transfers to human-annotated preference data.

\begin{table}[h]
\centering
\begin{threeparttable}
\resizebox{\columnwidth}{!}{%
\begin{tabular}{lcc}
\toprule
\textbf{Training Data} & \textbf{Selection Size} & \textbf{RewardBench Total} \\
\midrule
Ours & 10\% & 0.7052 \\
Random & 10\% & 0.6798 \\
Full Set & 100\% & 0.7108 \\
\bottomrule
\end{tabular}}
\end{threeparttable}
\caption{Reward model performance on the Anthropic Helpful-Harmless (HHH) human preference dataset. Our selected subset remains competitive with full-data training while outperforming a random subset of the same size.}
\label{tab:hhh_rewardbench}
\end{table}

\section{Further Analysis}
\label{app:analysis}

In this section, we provide a detailed analysis of our data selection method, exploring several key aspects and their impact on model performance. Specifically, we analyze the influence of different models for difficulty calculation, investigate the optimal selection ratio, and study the sensitivity of our method to response length. Additionally, we conducted a comprehensive statistical analysis on the data subset selected by our method.

\subsection{Impact of Different Models on Difficulty Calculation}

The calculation of the difficulty (i.e., DPO implicit reward gap) plays a central role in our data selection method. We explore how the choice of model for calculating the reward gap affects the selected subset of data and subsequent model performance.

\paragraph{Experimental Setup} We compare three different model pairs for calculating DPO implicit reward gaps: (1) \texttt{LLaMA3} series: LLaMA3-iterative-DPO-final and {LLaMA3-SFT} (our default setup), (2) \texttt{Gemma2} series: {Gemma-2-2b-it} and {Gemma-2-2b}, and (3) \texttt{Tulu3} series: {Tulu3-Llama3.1-8B-DPO} and {Tulu3-Llama3.1-8B-SFT}. For each model pair, we select 10\% of the \texttt{Skywork-Preference} dataset and train reward models using the same experimental protocol as described in Section~\ref{sec:experiments}.

\paragraph{Results} Table~\ref{tab:model_impact} presents the performance comparison across different selection models.

\begin{table}[h]
\centering
\begin{threeparttable}
\resizebox{\columnwidth}{!}{%
\begin{tabular}{cccccc}
\toprule
\textbf{Model} & \textbf{C} & \textbf{CH} & \textbf{S} & \textbf{R} & \textbf{Total} \\
\midrule
\texttt{LLaMA3} & 0.8798 & 0.7785 & 0.8446 & 0.6138 & 0.7588 \\
\texttt{Gemma2} & 0.8673 & 0.7739 & 0.8316 & 0.6143 & 0.7485 \\
\texttt{Tulu3} & 0.8692 & 0.7651 & 0.8476 & 0.6098 & 0.7502\\
\bottomrule
\end{tabular}}
\end{threeparttable}
\caption{Performance comparison using different model pairs for difficulty calculation on RewardBench. For each column, C refers to Chat part, CH refers to Chat-Hard part, S refers to Safety part, and R refers to Reasoning part. All methods select 10\% of the Skywork-Reward-Preference-80K-v0.2 dataset.}
\label{tab:model_impact}
\end{table}


Experimental results indicate that using different models to compute the DPO implicit reward gap does not significantly affect the quality of the selected data. This can be attributed to the fact that while the difficulty level of individual data points may vary across models, the ranking of these difficulties tends to remain consistent. In other words, data points that are considered difficult for one model are generally difficult for all models. This suggests that our approach is effective in identifying the challenging subset of the preference dataset, independent of the specific model choice.

\subsection{Investigation of the Optimal Selection Ratio}
\label{analysis:scaling-effects}

Understanding the relationship between subset size and model performance is crucial for the practical deployment of our method. We investigate how varying the proportion of selected data affects both reward model training and DPO fine-tuning performance.

\paragraph{Experimental Setup} We evaluate our method using different selection ratios: 5\%, 10\%, 15\%, 20\%, 30\%, and 50\% of the original \texttt{SHP} dataset. For each subset size, we train reward models and evaluate performance on RewardBench using the same experimental protocol as described in Section~\ref{sec:experiments}.

\begin{figure}[h]
\centering
\includegraphics[width=\linewidth]{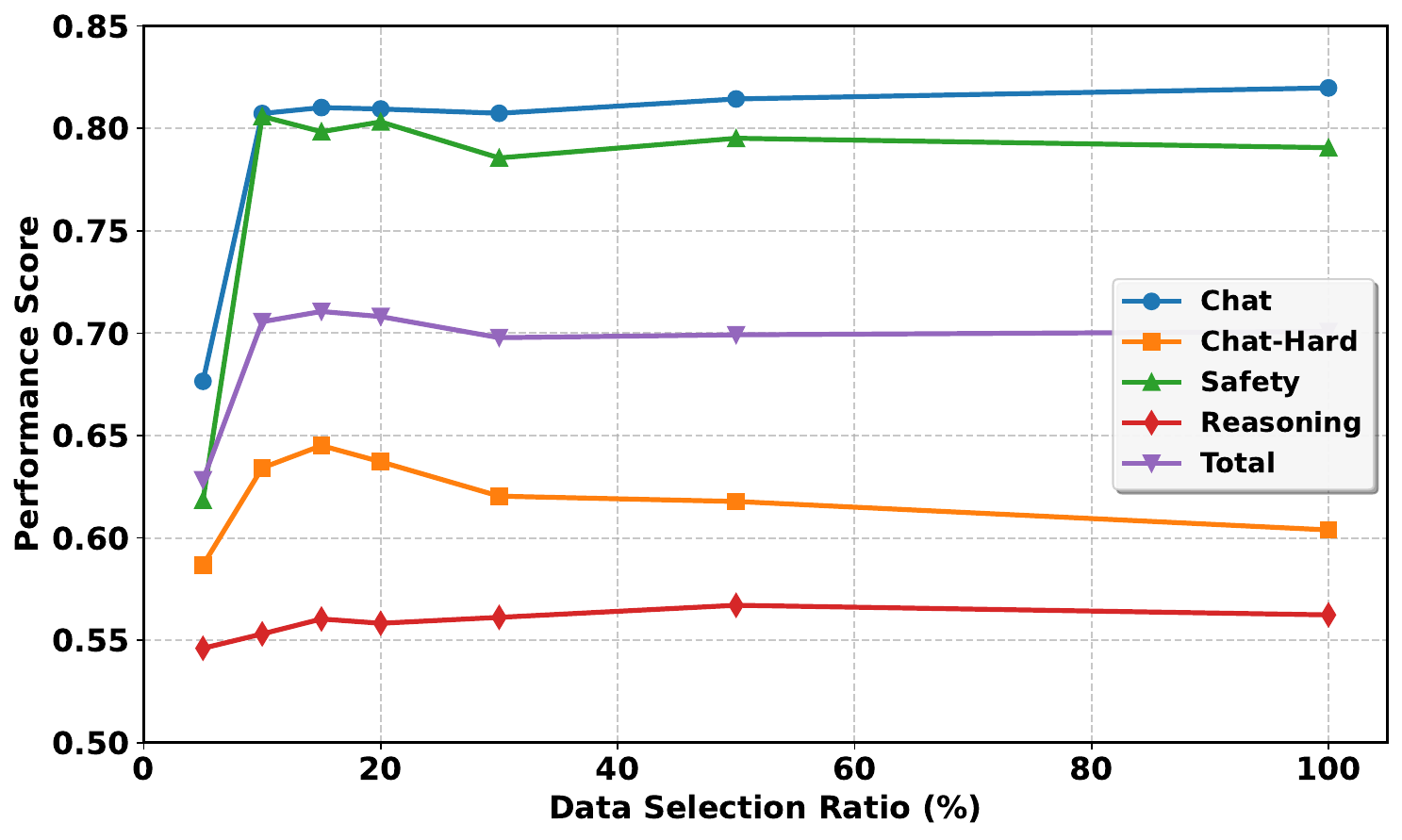}
\caption{Performance scaling effects with different data selection ratios on RewardBench using \texttt{SHP} dataset.}
\label{fig:scaling_effects}
\end{figure}

\paragraph{Results} Figure~\ref{fig:scaling_effects} and Table~\ref{tab:scaling_effects} show the performance trends across different data selection ratios.

\begin{table}[h]
\centering
\begin{threeparttable}
\resizebox{\columnwidth}{!}{%
\begin{tabular}{cccccc}
\toprule
\textbf{Ratio} & \textbf{C} & \textbf{CH} & \textbf{S} & \textbf{R} & \textbf{Total} \\
\midrule
5\% & 0.6765 & 0.5867 & 0.6184 & 0.5461 & 0.6283 \\
10\% & 0.8073 & 0.6342 & 0.8059 & 0.5531 & 0.7056 \\
15\% & 0.8102 & 0.6451 & 0.7984 & 0.5604 & 0.7106\\
20\% & 0.8095 & 0.6372 & 0.8032 & 0.5583 & 0.7081\\
30\% & 0.8074 & 0.6204 & 0.7856 & 0.5612 & 0.6978\\
50\% & 0.8144 & 0.6178 & 0.7952 & 0.5671 & 0.6992\\
100\% & 0.8198 & 0.6039 & 0.7906 & 0.5624 & 0.7008\\
\bottomrule
\end{tabular}}
\end{threeparttable}
\caption{Performance scaling effects with different data selection ratios on RewardBench using \texttt{SHP}  dataset. For each column, C refers to Chat part, CH refers to Chat-Hard part, S refers to Safety part, and R refers to Reasoning part. (This is an alternative illustration of Figure \ref{fig:scaling_effects}.)}
\label{tab:scaling_effects}
\end{table}

The results demonstrate diminishing returns as the selection ratio increases beyond 10-15\%. This finding suggests that our method effectively identifies the most valuable examples within a relatively small subset, with additional data providing marginal improvements. The optimal selection ratio appears to be around 10-15\%, balancing data efficiency with performance gains. When the proportion of selected data exceeds 20\%, the performance improvement becomes less pronounced. Further increasing the selection ratio may lead to a decrease in training efficiency.

\subsection{Impact of Response Length on Data Selection}
\label{analysis:length-impact}

A potential concern with our method is whether the cumulative nature of DPO reward calculation introduces bias toward longer responses. We investigate the impact of length normalization on our selection method to understand whether raw reward gaps or length-normalized gaps lead to better data selection.

\paragraph{Experimental Setup} We compare two variants of our difficulty calculation: (1) raw DPO implicit reward gap without normalization, and (2) length-normalized DPO implicit reward gap. The two approaches are formally defined as:

\begin{definition}[Raw Reward Gap]
\begin{equation}
\Delta r_{\text{raw}} \triangleq r_{\text{DPO}}(x, y_{w}) - r_{\text{DPO}}(x, y_{l}).
\end{equation}
\end{definition}

\begin{definition}[Length-Normalized Reward Gap]
\begin{equation}
\Delta r_{\text{norm}} \triangleq \frac{r_{\text{DPO}}(x, y_{w})}{|y_{w}|} - \frac{r_{\text{DPO}}(x, y_{l})}{|y_{l}|},
\end{equation}
where $|y|$ denotes the token length of response $y$.
\end{definition}
We select 10\% of the Skywork-Preference dataset using both methods and evaluate the resulting reward models on RewardBench.

\paragraph{Results} Table~\ref{tab:length_normalization} presents the performance comparison between raw and length-normalized reward gap calculations.

\begin{table}[h]
\centering
\begin{threeparttable}
\resizebox{\columnwidth}{!}{%
\begin{tabular}{cccccc}
\toprule
\textbf{Method} & \textbf{C} & \textbf{CH} & \textbf{S} & \textbf{R} & \textbf{Total} \\
\midrule
Raw & 0.8798 & 0.7785 & 0.8446 & 0.6138 & 0.7588  \\
L-N & 0.8692 & 0.7590 & 0.8267 & 0.6074 & 0.7416 \\
\bottomrule
\end{tabular}}
\end{threeparttable}
\caption{Performance comparison between raw and length-normalized reward gap calculations on RewardBench using \texttt{Skywork-Preference} dataset. Raw refers to the raw DPO implicit reward gap, and L-N refers to the length-normalized DPO implicit reward gap defined above. For each column, C refers to Chat part, CH refers to Chat-Hard part, S refers to Safety part, and R refers to Reasoning part.}
\label{tab:length_normalization}
\end{table}

Experimental results show that normalizing the response length when computing the DPO implicit reward gap does not improve the quality of the selected data. This may be due to the fact that longer responses inherently provide more reward signals, which could aid the model in learning more effectively. Therefore, normalizing the response length might not be appropriate, as it could result in the selection of data points with shorter responses where individual chosen response tokens have low generation probabilities (or rejected response tokens have high generation probabilities). However, these data points are unlikely to contribute significantly to the model's performance improvement.

To directly test whether our gains are explained by response length alone, we also compare against a length-only baseline that selects the top 10\% longest pairs from \texttt{Skywork-Preference} and uses the same reward model training and evaluation protocol. As shown in Table~\ref{tab:length_only_baseline}, the length-only subset performs substantially below our reward-gap selection, suggesting that our method does not simply select longer responses.

\begin{table}[h]
\centering
\begin{threeparttable}
\resizebox{\columnwidth}{!}{%
\begin{tabular}{lcc}
\toprule
\textbf{Selection Rule} & \textbf{Training Data} & \textbf{RewardBench Total} \\
\midrule
Ours & 10\% & 0.7588 \\
Length-only & 10\% & 0.7302 \\
Random & 10\% & 0.7189 \\
Full Set & 100\% & 0.7812 \\
\bottomrule
\end{tabular}}
\end{threeparttable}
\caption{Length-only baseline on \texttt{Skywork-Preference}. Selecting the longest 10\% pairs improves over random selection but remains clearly below our reward-gap-based selection.}
\label{tab:length_only_baseline}
\end{table}




\subsection{Analysis of Selected Data Examples}

To provide a more intuitive understanding of our selected data, we present several statistical characteristics of the data filtered by our method. These statistical features are derived from 10\% of the data filtered from the \texttt{Skywork-Preference} dataset.

\begin{figure}[h]
    \centering
    \includegraphics[width=\linewidth]{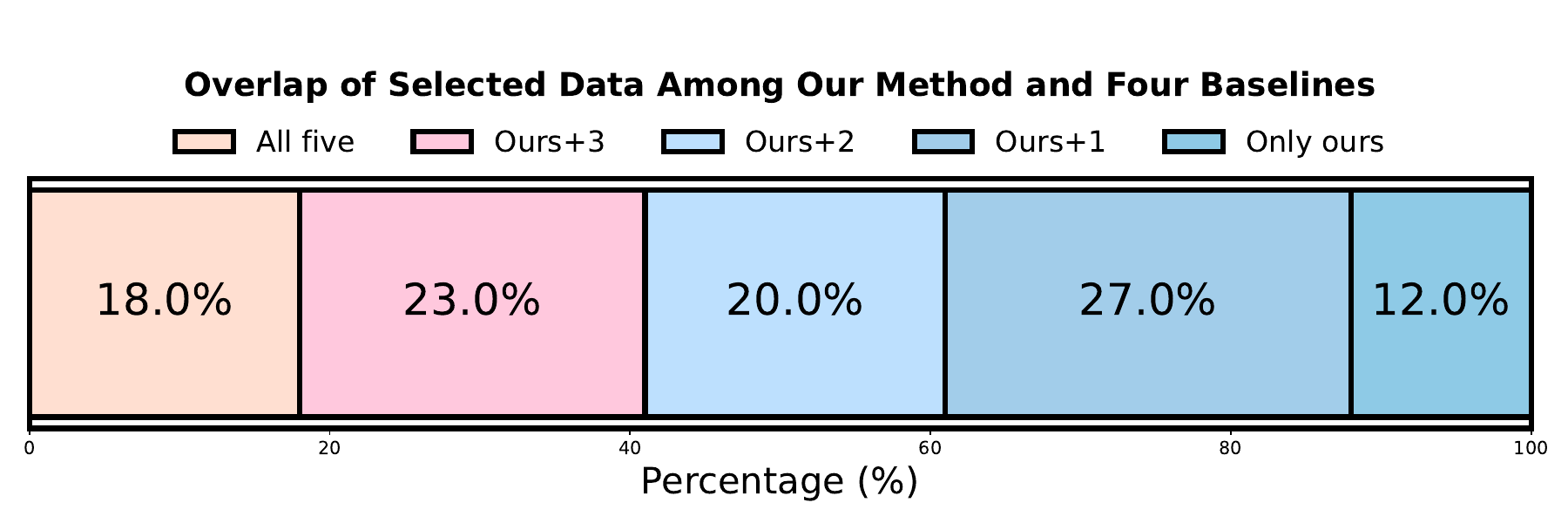}
    \caption{
        The overlap of selected data among our method and four baselines.
        The legend indicates selection agreement: \textbf{All five} indicates that the data is chosen by all methods. \textbf{Ours+3} indicates that the data is chosen by our method and three baselines, and so on. \textbf{Only ours} indicates that the data is only chosen by our approach.
    }
    \label{fig:data_overlap}
\end{figure}

\begin{table}[htbp]
\centering
\begin{threeparttable}
\resizebox{\columnwidth}{!}{%
\begin{tabular}{lcc}
\toprule
\textbf{\textbf{Data Subset}} & \textbf{Avg. Tokens (W)} & \textbf{Avg. Tokens (L)} \\
\midrule
Original Dataset & 2057& 2337\\
Our Method & 2198 & 2506\\
Unique to Our & 2314& 2618\\
\bottomrule
\end{tabular}}
\end{threeparttable}
\caption{Average token length of responses in different data subsets. \textbf{Unique to Our} refers to the subset only selected by our method compared to other baselines.}
\label{tab:length_stats}
\end{table}

\paragraph{Overlap with Data Selected by Other Methods}

Figure~\ref{fig:data_overlap} illustrates the overlap between data selected by our method and that selected by baseline methods. As shown, our approach identifies a substantial proportion of unique data points that are not captured by alternative filtering techniques.

\paragraph{Length Characteristics of Selected Data}
Given our discussion on the impact of response length on training effectiveness, we analyze the length characteristics of the data selected by our method.

Table~\ref{tab:length_stats} demonstrates that the data filtered by our method has a significantly higher average length compared to the overall dataset average. This indicates that our approach tends to select longer responses, which potentially carry more reward signals, thereby contributing to improved training outcomes.

\end{document}